\documentclass[lettersize,journal]{IEEEtran}
\usepackage{amsmath,amsfonts}
\usepackage{algorithmic}
\usepackage{algorithm}
\usepackage{array}
\usepackage[caption=false,font=normalsize,labelfont=sf,textfont=sf]{subfig}
\usepackage{textcomp}
\usepackage{amsmath,amssymb,amsfonts}
\usepackage{algorithmic}
\usepackage{graphicx}
\usepackage{textcomp}
\usepackage{amsmath,amssymb,amsfonts}
\usepackage{algorithmic}
\usepackage{graphicx}
\usepackage{textcomp}
\usepackage{wrapfig}
\usepackage{amsmath,amssymb,amsfonts}
\usepackage{algorithmic}
\usepackage{graphicx}
\usepackage{textcomp}
\usepackage{array}
\DeclareMathOperator{\E}{\mathbb{E}}
\usepackage[caption=false,font=normalsize,labelfont=sf,textfont=sf]{subfig}
\usepackage{stfloats}
\usepackage{url}
\usepackage{verbatim}
\usepackage{graphicx}
\usepackage{color}
\usepackage{cite}
\usepackage{multirow}
\usepackage{makecell}
\usepackage{amsmath}
\usepackage{amssymb}
\usepackage[frak=mma]{mathalfa}
\usepackage{bbm}
\usepackage[mathscr]{euscript}
\usepackage{stfloats}
\usepackage{url}
\usepackage{verbatim}
\usepackage{graphicx}
\usepackage{cite}
\begin{document}

\title{Task-Decoupled Image Inpainting Framework for Class-specific Object Remover}

\author{Changsuk Oh and H. Jin Kim
        % <-this % stops a space
\thanks{Changsuk Oh and H. Jin Kim with the Department of Aerospace Engineering, Seoul National University, Seoul, Korea (email: santgo@snu.ac.kr and hjinkim@snu.ac.kr)}% <-this % stops a space
}

\markboth{Journal of \LaTeX\ Class Files,~Vol.~14, No.~8, August~2021}%
{Shell \MakeLowercase{\textit{et al.}}: A Sample Article Using IEEEtran.cls for IEEE Journals}


\maketitle

\begin{abstract}
Object removal refers to the process of erasing designated objects from an image while preserving the overall appearance. Existing works on object removal erase removal targets using image inpainting networks. However, image inpainting networks often generate unsatisfactory removal results. In this work, we find that the current training approach which encourages a \textit{single} image inpainting model to handle \textit{both} object removal and restoration tasks is one of the reasons behind such unsatisfactory result. Based on this finding, we propose a task-decoupled image inpainting framework which generates two separate inpainting models: an object restorer for object restoration tasks and an object remover for object removal tasks. We train the object restorer with the masks that partially cover the removal targets. Then, the proposed framework makes an object restorer to generate a guidance for training the object remover. Using the proposed framework, we obtain a class-specific object remover which focuses on removing objects of a target class, aiming to better erase target class objects than general object removers. We also introduce a data curation method that encompasses the image selection and mask generation approaches used to produce training data for the proposed class-specific object remover. Using the proposed curation method, we can simulate the scenarios where an object remover is trained on the data with object removal ground truth images. Experiments on multiple datasets show that the proposed class-specific object remover can better remove target class objects than object removers based on image inpainting networks. 
\end{abstract}

\section{Introduction}

\begin{figure}[t!]
  \centering
  \includegraphics[width=0.45\textwidth]{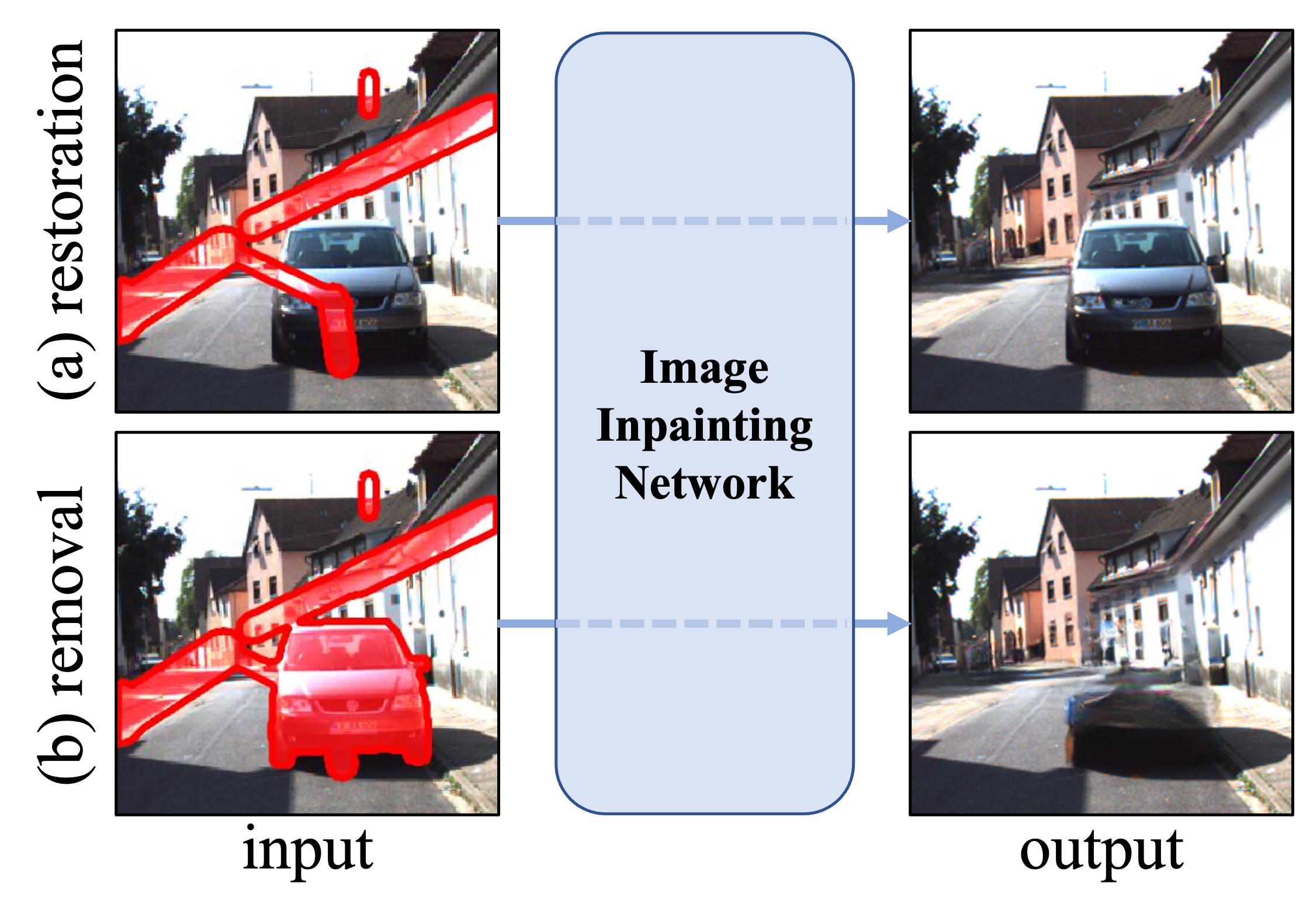}
  \caption{Two tasks of an image inpainting network. Lama \cite{suvorov2022resolution} is utilized for inpainting.} 
    \label{fig:thumbnail}
\end{figure}

Object removal aims to plausibly remove unwanted objects in an image. Recently, \cite{liu2018image, suvorov2022resolution, zeng2022aggregated,cao2021learning, rombach2022high,yi2020contextual, zeng2020high, 9775085, 9751095} generate object removal results using their image inpainting networks. The image inpainting networks can perform object removal because they are exposed to both object removal and restoration tasks during training. The training method of current image inpainting networks \cite{yi2020contextual, zeng2020high, he2022masked,zeng2021cr, 9113276, 9813740, 10445705, 10068561} makes training samples by superimposing randomly sampled masks on intact images, prompting the inpainting networks to restore or remove an object based on the degree of occlusion caused by a randomly sampled mask. Specifically, as shown in Fig. \ref{fig:thumbnail}, if an object (car) is partially occluded by a mask (Fig. \ref{fig:thumbnail}(a)), an image inpainting network attempts to plausibly restore the object’s appearance based on the remaining visible areas. On the other hand, when a mask covers an object entirely (Fig. \ref{fig:thumbnail}(b)), the inpainted image may not include the object because there is no information about the object in the unmasked region. 

Although image inpainting networks learn how to remove objects in an image during training, they often generate unsatisfactory object removal results, which can be seen in Fig. \ref{fig:thumbnail}(b), Fig. \ref{fig:qualitative_car}, and Fig. \ref{fig:qualitative_human}. Our studies find that the current training approach which encourages a single inpainting network to handle both object removal and restoration tasks is one of the reasons behind such unsatisfactory result. The objective of the current inpainting networks is to inpaint the missing areas to be (perceptually) similar to the original image. Therefore, it is suitable for training the restoration task. However, for an object removal task, the inpainted area where the removal target was located should be filled with visual features that harmonize well with the background, which is not (perceptually) similar to the corresponding area of the original image. This implies that the loss function of current image inpainting networks, which utilizes an original image for output quality evaluation, cannot properly train an object remover.

Motivated by the observation, we propose a task-decoupled image inpainting framework which generates two separate models: an object restorer for object restoration tasks and an object remover for object removal tasks. We train an image inpainting network as an object restorer with the masks that partially cover restoration targets. Then, the proposed framework leverages the knowledge from the restorer to train the object remover. The restorer provides informative guidance on which visual features  the remover should avoid when generating its output and which areas of an object restoration result that the discriminator (of the remover) should not consider real. 

Using the proposed framework, we develop a class-specific object remover which specializes in removing objects of a target class. Developing a class-specific model \cite{singh2019finegan, hinz2018generating, yang2017lr, li2021collaging, tang2020local, huang2017beyond, gu2019mask, li2018global}, focusing on one or selected classes, is a common approach in image generation to improve the output quality. We define a class-specific object remover as a tool designed to effectively handle masked images, where removal targets belong to one class, and only the pixels corresponding to the removal targets are masked. 

Developing a class-specific object remover can contribute to obtaining high-quality results in real-world object removal scenarios. Specifically, when a user designates removal targets by clicking a pixel of each target or specifies them through human language, we need a segmentation model to identify pixels corresponding to the removal targets. In this scenario, where a segmentation model generates a mask that tightly covers removal targets, a class-specific object remover designed to effectively handle such masked images can help generating high-quality target class object removal results. 

The proposed framework first trains an image inpainter as a class-specific object restorer using masks that partially cover target class objects in images. As the restoration task is suitable for using an original image as a reference for output quality evaluation, the proposed framework employs the current image inpainting model \cite{suvorov2022resolution} for training. To obtain a class-specific object remover, the proposed framework utilizes not only the guidance from a class-specific object restorer but also a data curation method. The proposed data curation method generates input data by covering images without a target class object with masks of target class object shape. We refer to masks of target class object shape as class-shaped masks in this paper. By employing the curated data, we can simulate the scenarios where a class-specific object remover is trained on data with class-wise object removal ground truth (GT), allowing the class-specific remover to handle various class-shaped masks during training. Class-wise object removal is the task where all objects in an image belonging to a target class are designated as removal targets. We test our method on COCO \cite{lin2014microsoft}, RORD \cite{Sagong_2022_BMVC}, the autonomous vehicle datasets (KITTI \cite{Geiger2012CVPR}, STEP \cite{Weber2021NEURIPSDATA}, KITTI-360 \cite{Liao2022PAMI}, Mapillar \cite{Neuhold_2017_ICCV}, and Cityscape \cite{Cordts_2016_CVPR}), and validate that the proposed model can better erase target class objects than the image inpainting networks. 

This paper has the following contributions:
\begin{itemize}
    \item We look closely into the reason behind the unsatisfactory object removal results made by image inpainting networks and find that the current training approach which encourages a single inpainting model to handle \textit{both} object removal and restoration tasks can be one of the reasons. 
    \item To tackle the above problem, we propose a task-decoupled image inpainting framework which generates two separate inpainting models: an object restorer for object restoration tasks and an object remover for object removal tasks.
    \item Using the proposed framework, we develop a class-specific object remover, which is designed to better remove target class objects than object removers based on image inpainting networks in real-world object removal scenarios.
    \item Experiments on the multiple image datasets demonstrate that the class-specific object removers obtained by using the proposed task-decoupled inpainting framework can better remove target class objects compared to the object removers based on image inpainting networks.
\end{itemize}

\section{related works}
\subsection{Image Inpainting Network Training}
The current training framework for a data-driven image inpainting network generates training data by superimposing synthetic masks on intact images and evaluates the quality of intermediate output by comparing it with the original image. Rectangular-shaped patches \cite{yu2018generative, iizuka2017globally, Yan_2018_ECCV, Yang_2017_CVPR, wang2019laplacian, zhang2022gan}, irregular-shaped patches \cite{Yu_2021_ICCV, li2020deepgin, wang2020vcnet, shin2020pepsi++, quan2022image, liu2022reduce,liu2021pd, 9775085, 9751095, 9113276, 9813740, 10068561, 10445705}, and object-shaped patches \cite{rombach2022high, yi2020contextual, cao2021learning} are placed at random locations to generate training data. Then, L1 \cite{zeng2021cr,he2022masked, yi2020contextual, zeng2020high, Wang_2022_CVPR, liu2021pd, yu2019free}, L2 \cite{lugmayr2022repaint, rombach2022high}, and variants of L1 \cite{wang2018image, suin2021distillation} functions are exploited to compute the pixel-level difference between a completed image and the original image. \cite{suvorov2022resolution,zheng2022image,cao2022learning, dong2022incremental} and \cite{li2022mat,jam2021r} utilize HRFPL \cite{suvorov2022resolution} and VGG-based perceptual loss \cite{johnson2016perceptual} functions to ensure that the inpainted images are perceptually similar to the original images. And \cite{zhu2021image,yu2022high,shetty2018adversarial,cao2021learning,guo2021image} use both pixel-level reconstruction losses and perceptual losses for training. 

Since the randomly sampled masks can cover an object partially or entirely, using the current training framework, an inpainter encounters both restoration and removal tasks during training. In the restoration task, an original image can be utilized as a reference for evaluating the quality of the inpainted image. On the other hand, in the removal task, it is inappropriate to use an original image for evaluating the output quality because removal targets exist in the original image. However, the current training framework always utilizes the original image as a reference to evaluate the output quality, which can be one major reason that image inpainting networks generate unsatisfactory object removal results. 

To obtain an image inpainter with superior removal performance, we design a task-decoupled image inpainting framework. The framework generates two separate models: an object restorer for object restoration and an object remover for object removal. As we can utilize original images as GT for training object restoration tasks, the current inpainting model is utilized to obtain an object restorer. Then, for an object remover, we use the guidance generated by the restorer, as original images are not suitable to be used as reference.

\begin{figure}[t!]
  \centering
  \includegraphics[width=0.45
  \textwidth]{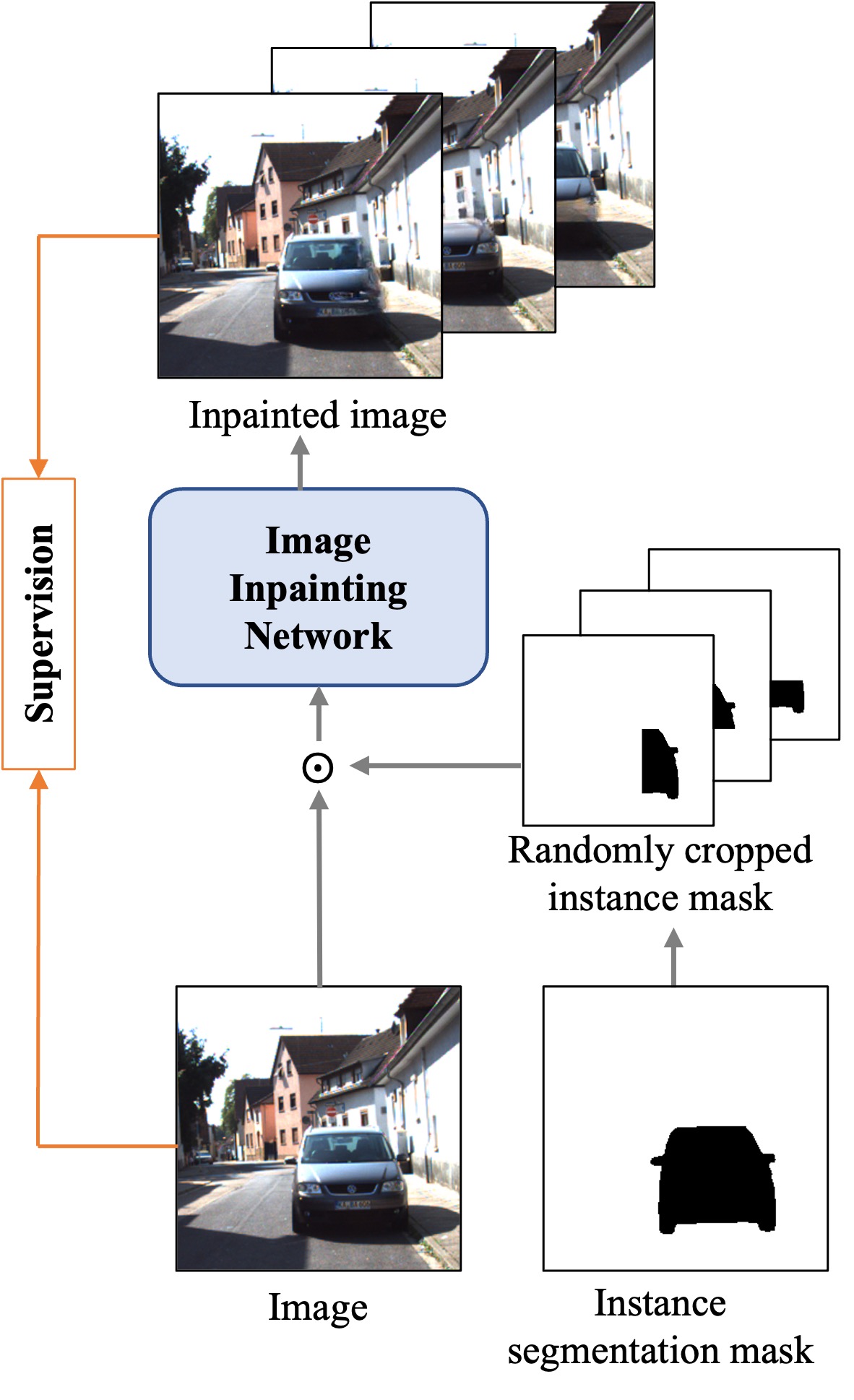}
  \caption{The class-specific object restorer training process.}
    \label{fig:restorer}
\end{figure}

\subsection{Class-specific Image Generation}
Developing a class-specific model dedicated to specific classes is a common strategy \cite{singh2019finegan,hinz2018generating,yang2017lr,li2021collaging,tang2020local,huang2017beyond, gu2019mask, li2018global} in image generation to enhance the output quality. \cite{singh2019finegan,hinz2018generating,yang2017lr} utilize two models to generate an image, with one dedicated to foreground objects and the other for the background. \cite{li2021collaging,tang2020local} construct generators for each class and synthesize a semantic-guided image by utilizing them. \cite{huang2017beyond, gu2019mask, li2018global} produce realistic facial images by obtaining models that specialize in generating crucial parts, such as the eyes and mouth. 

In this paper, we develop a class-specific object remover using the proposed task-decoupled image inpainting framework. The proposed framework utilizes data curation to gather training data that can simulate the scenarios where an object remover is trained on the data with class-wise object removal GT. This ensures that the object remover handles various class-shaped masks during training, which can help the remover to learn class-specific information for removal during training. Furthermore, by leveraging guidance made by a class-specific object restorer, the proposed framework induces a class-specific object remover to refrain from generating visual features similar to the appearance of target class objects when filling the masks of target class object shape.

\section{Method}
In this section, we introduce a task-decoupled image inpainting framework utilized to develop a class-specific object remover.

\begin{figure*}[t!]
  \centering
  \includegraphics[width=1.0\textwidth]{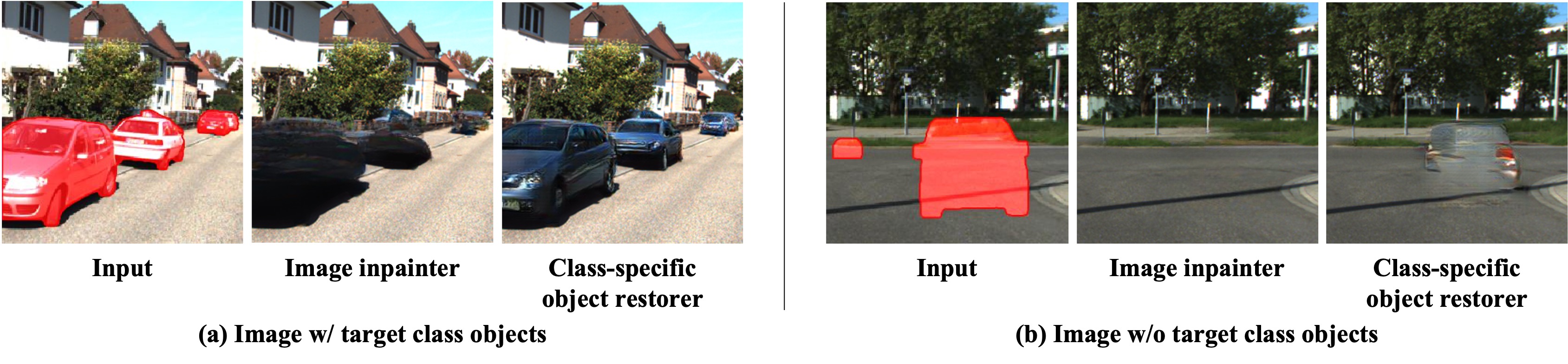}
  \caption{Inpainting results of the inpainting network (Lama \cite{suvorov2022resolution}) and the proposed class-specific object restorer. \textit{Car} class is set as a target class.}
    \label{fig:restorer_result}
\end{figure*}

\subsection{Class-specific Object Restorer}
Object restoration is a task where an original image can be used as a reference for evaluating the output quality. Therefore, the proposed framework employs the current image inpainting model for training. To train an image inpainting model as a class-specific object restorer, we need data where target class objects are partially occluded. We select images where target class objects occupy 5--40\% of the pixels. When target class objects are too large or too small, it is challenging for a restorer to learn class-specific features such as the overall shape and visual features of the target class objects. We generate masks that cover each restoration target by 40-60\% using the instance segmentation maps. Then, the image inpainting network is trained to restore the partially covered target class objects in a plausible way, as shown in Fig. \ref{fig:restorer}.

The object restoration result using the proposed class-specific object restorer is presented in Fig. \ref{fig:restorer_result}(a). We can observe that the class-specific object restorer plausibly completes the masked region with car-like visual features, while the image inpainting model generates dark traces of the removal targets. One interesting feature of the class-specific object restorer is demonstrated in Fig. \ref{fig:restorer_result}(b). The class-specific object restorer generates visual features similar to the target class objects even when no target class objects are present in the original image and only a class-shaped mask is provided for restoration. This result demonstrates that a class-specific object restorer utilizes class-specific information for inpainting, regardless of the presence of target class objects in an original image, as long as the image is covered by a class-shaped mask.

\subsection{Data Curation for Class-specific Object Remover}

To train an image inpainter as a class-specific object remover, we need masked images where target class objects are covered by class-shaped masks and reference images for evaluating the quality of intermediate outputs (inpainted images). Unlike object restoration, we cannot utilize an original image as a reference in object removal, as the original image contains removal targets. Therefore, we need the object removal GT images taken under the identical environment of input images, only without the removal targets. However, obtaining such data is challenging. Thus, we introduce a data curation method designed to create training data that simulates scenarios where an inpainter is trained on data with object removal GT images. 

The proposed data curation method selects images without target class objects and utilizes class-shaped masks to generate training data. By doing so, we can obtain curated data that an object remover cannot distinguish from the input images of class-wise object removal tasks. As a class-wise object removal aims to erase all target class objects in an image, by using input images of class-wise object removal tasks for training, we can make a class-specific object remover to inpaint various class-shaped masks and to learn class-specific information for removal during training. 

\begin{figure}[t]
  \centering
  \includegraphics[width=0.45\textwidth]{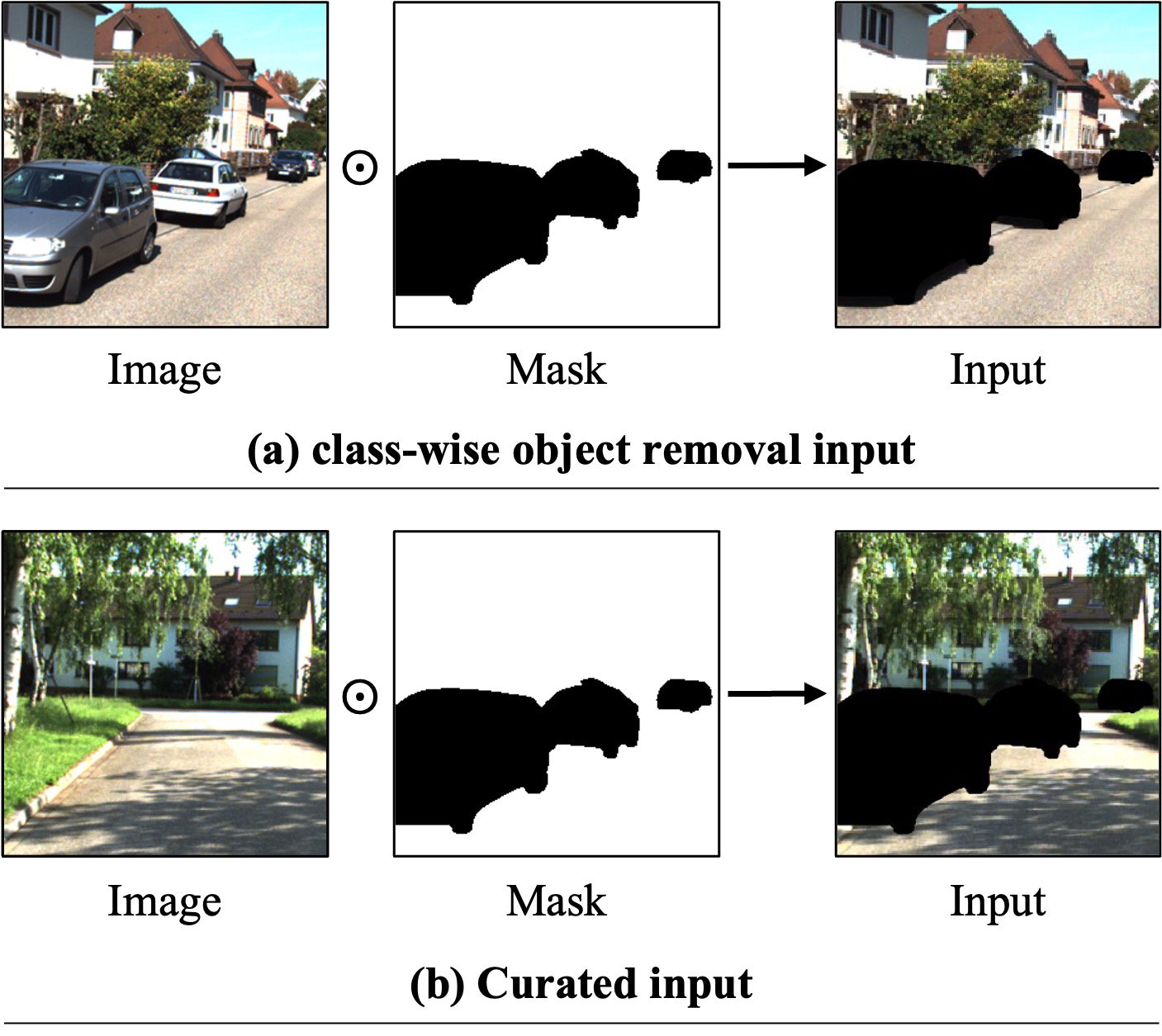}
  \caption{Input masked images. (a) shows an input image of a class-wise object removal task. (b) demonstrates a masked image generated using the proposed data curation method.}
    \label{fig:curation}
\end{figure}

The masked image generated by the proposed data curation method is equivalent to the input image (masked image) of the class-wise object removal task in two aspects, which can be seen in Fig. \ref{fig:curation}. First, it is occluded by the class-shaped mask. Second, there are no target class objects in the unmasked region. Therefore, from the perspective of an image inpainter, completing the masked image generated by the proposed data curation method is very similar to completing the masked image of a class-wise object removal task. However, there is one significant difference between the two completing tasks. In contrast to an original image of a class-wise object removal task that contains removal targets, the original image of the training data generated by the data curation method does not contain target class objects. This implies that the original image of the curated data is suitable to be used as a reference for supervision. We also use randomly sampled masks \cite{suvorov2022resolution} to ensure the overall inpainting performance.

\subsection{Guidance}
A class-specific object restoration result is a good example that shows what the class-specific object remover should avoid when completing a class-shaped mask. Based on this observation, we design $\mathscr{L}_{afterimage}$, which is tailored to encourage the output of the class-specific object remover to be dissimilar to that of the class-specific object restorer. $\mathscr{L}_{afterimage}$ uses the high receptive field model ($\phi_{HRF}$) \cite{suvorov2022resolution, dong2022incremental, cao2022learning} to calculate the perceptual difference between the two inpainting outputs as follows:
\begin{align}
   \mathscr{L}_{afterimage} = -\mathcal{M}([\phi_{HRF}(\hat{I}) - \phi_{HRF}(\hat{I}_{rest})]^2) ,
\end{align} 
where $\hat{I} = G(I\odot m)$ and $\hat{I}_{rest} = G_{rest}(I \odot m)$. $G$ and $G_{rest}$ are the generator of the class-specific object remover and restorer, respectively. $I$ is an original image and $m$ indicates an input mask whose pixel values are equal to one, except for the pixels from removal targets, which have zero intensity. $\mathcal{M}$ stands for the sequential two-stage mean operation \cite{suvorov2022resolution}. 

Additionally, the output of the class-specific object restorer is also utilized to improve the performance of the class-specific object remover's discriminator ($D$). Our adversarial loss term uses the additional sample as follows:
\begin{align}
\centering
    \mathscr{L}_{adv} =   \mathscr{L}_G + \mathscr{L}_D ,
\end{align} 
where
\begin{align}
    \mathscr{L}_{G} = & -\mathbb{E}_{I,m} \left[ \log D(\hat{I})\right], \notag \\
    \mathscr{L}_{D} = & -\mathbb{E}_{I} \left[ \log D(I)\right] -\mathbb{E}_{I,m} [ \log D(\hat{I}) \odot  m   + \\ \notag
    & \log \{(1- D(\hat{I}))(1- D(\hat{I}_{rest})) \} \odot (\mathbbm{1}_{m}-m) ]. 
\end{align} 
$\mathbbm{1}_{m}$ is a binary mask with the same size as the mask $m$ and all pixel values equal to one. 

\begin{figure*}[t!]
  \centering
  \includegraphics[width=0.85\textwidth]{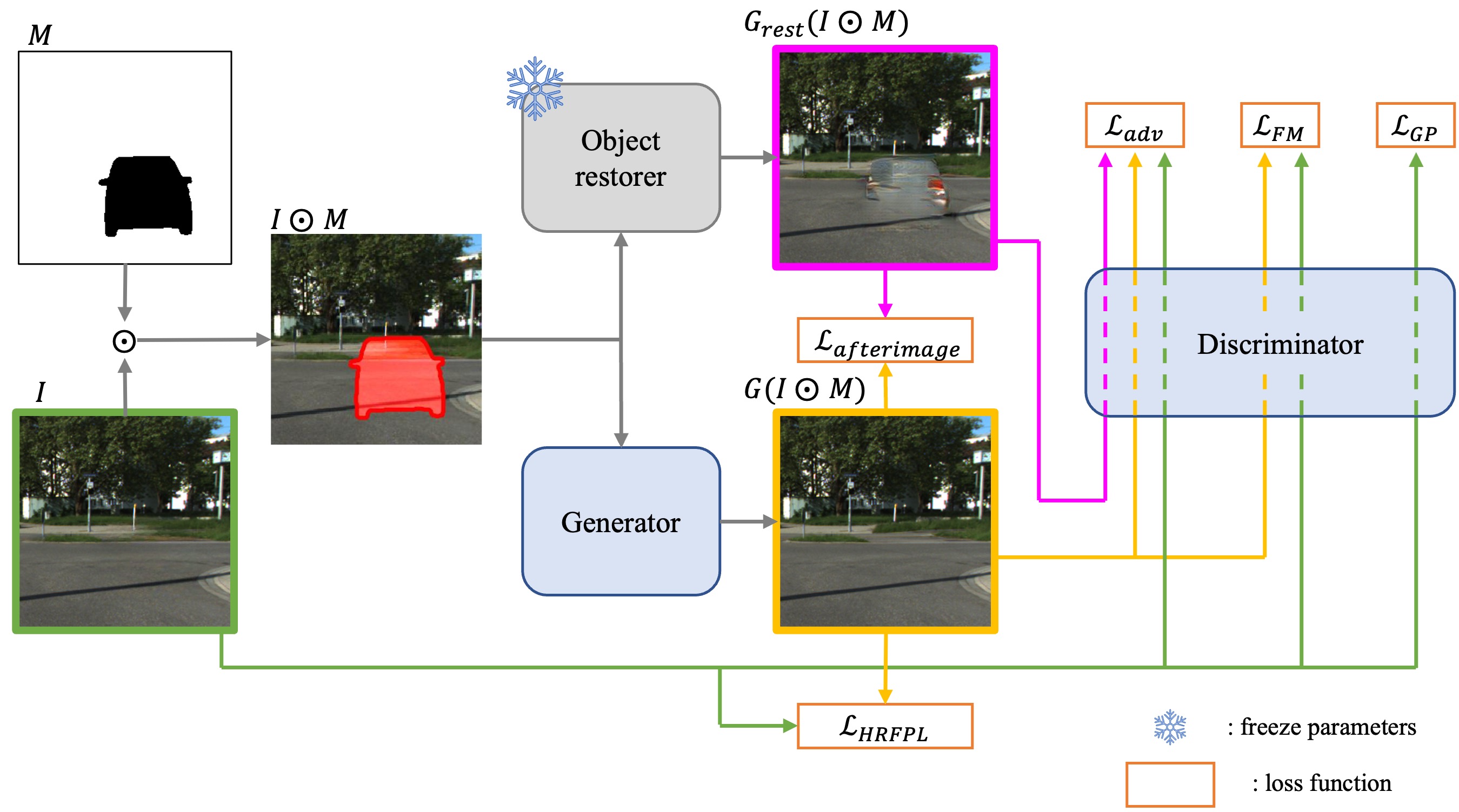}
  \caption{Class-specific object remover training process.}
    \label{fig:method}
\end{figure*}

The proposed framework employs the generator structure of Lama \cite{suvorov2022resolution} that demonstrates decent image inpainting results using fast Fourier convolutions (FFCs) \cite{suvorov2022resolution, zheng2022image}. We utilize a patch-level discriminator \cite{Isola_2017_CVPR}. The proposed framework does not have additional computation during inference as the additional model is only used for training. For the overall loss function, we also use two perceptual loss functions ($\mathscr{L}_{HRFPL}$ \cite{suvorov2022resolution} and $\mathscr{L}_{FM}$ \cite{rombach2022high}). $\mathscr{L}_{HRFPL}$ evaluates the perceptual difference between an original image and an inpainted image using the high receptive field model ($\phi_{HRF}$) \cite{suvorov2022resolution, dong2022incremental, cao2022learning}. 
\begin{align}
   \mathscr{L}_{HRFPL} = \mathcal{M}([\phi_{HRF}(I) - \phi_{HRF}(\hat{I})]^2) .
\end{align} 
$\mathscr{L}_{FM}$ is designed to reduce the difference between the feature vectors of an original image and an inpainted image by using the discriminator as a feature extractor. We calculate $\mathscr{L}_{FM}$ as follows:
\begin{align}
   \mathscr{L}_{FM} = \frac{1}{T}\sum_{i=1}^{T} \frac{\mathcal{L}_{2} (D_{i}(I) , D_{i}(\hat{I}) )}{N_{i}} ,
\end{align} 
where T is total number of layers and $N_{i}$ indicates the number of elements of each layer of the discriminator. $\mathscr{L}_{GP}$ penalizes the gradient of the discriminator on real data as follows:
\begin{align}
    \mathscr{L}_{GP} = \E_{I} [ || \nabla D(I) ||^2  ].
\end{align}
The overall loss function can be summarized as
\begin{align}
\centering 
    \mathscr{L} = & \lambda_{AI} \mathscr{L}_{afterimage} + \lambda_{adv} \mathscr{L}_{adv} + \lambda_{PL}  \mathscr{L}_{HRFPL} \\ \notag
    + & \lambda_{FM} \mathscr{L}_{FM} + \lambda_{GP} \mathscr{L}_{GP},
\end{align}
where the hyper-parameters are empirically set as $\lambda_{AI}$ = 7, $\lambda_{adv}$ = 10, $\lambda_{PL}$ = 30, $\lambda_{FM}$ = 100, and $\lambda_{GP}$ = 0.001.

\section{Experiments and Results}
\textbf{Datasets. }
We conduct experiments using two datasets. First, we use the COCO dataset \cite{lin2014microsoft} and set \textit{person} as the target class. The class-specific object restorer and remover are trained using the COCO train set images with and without target class objects, respectively. We use the COCO validation set images as a test set.

Second, we create an assorted vehicle dataset by collecting data from the KITTI \cite{Geiger2012CVPR}, STEP \cite{Weber2021NEURIPSDATA}, KITTI-360 \cite{Liao2022PAMI}, Mapillar \cite{Neuhold_2017_ICCV}, and Cityscape \cite{Cordts_2016_CVPR} datasets. In this case, we set \textit{car} as a target class. We divide train set images of the assorted vehicle dataset into images with and without target class objects, and use them to train the restorer and remover, respectively. We use the KITTI object detection dataset \cite{Geiger2012CVPR} as a test set, while the remaining datasets are used as train and validation sets. The images from the KITTI-360 and STEP datasets are divided into four parts in the width direction. The images from the Cityscape and Mapillary datasets are resized to 1024 $\times$ 512 and 512 $\times$ 512, and horizontally divided into four equal parts (Cityscape) and two equal parts (Mapillary), respectively. Then, each image is center-cropped to 256 $\times$ 256. The class list included in the car category and the number of data for each dataset are presented in Table \ref{table:data}. To simulate scenarios where the removal targets are designated using a segmentation model, we obtain input masks for performance comparison using a semantic segmentation model (MSeg \cite{lambert2020mseg}). 

For the performance comparison, similar to \cite{cao2021learning}, we only use images where the removal targets cover 5--40\% of the images. For training class-specific object remover, class-shaped masks are obtained using semantic segmentation annotations, and a mask is randomly selected among the class-shaped masks to cover an image without a target class object. 

\begin{figure*}[th!]
  \centering
  \includegraphics[width=0.98\textwidth]{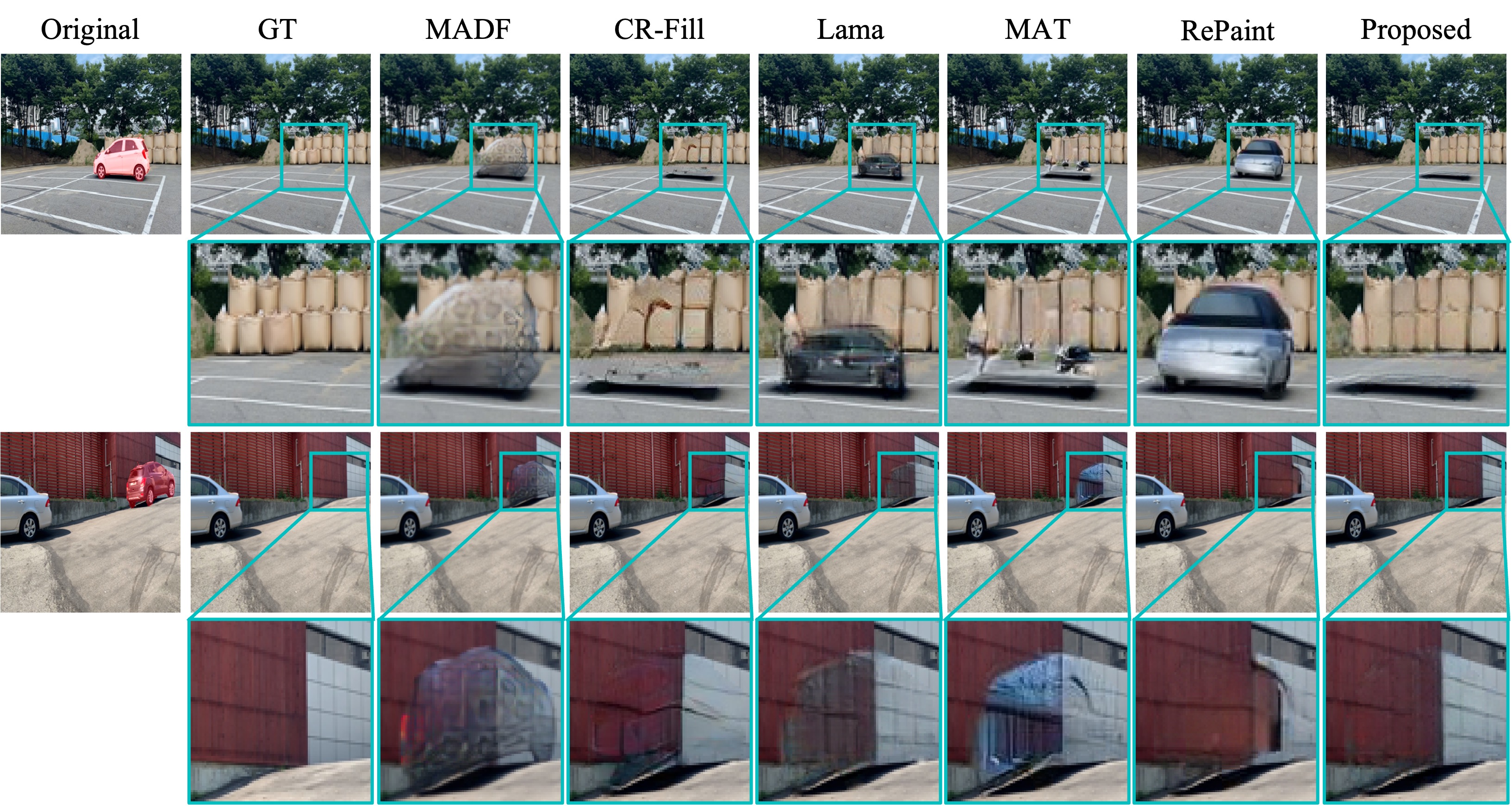}
  \caption{Car removal results on the RORD dataset. We use the ground truth semantic segmentation maps to generate input masks and use the models trained on PLACES2 to generate the results.
  }
    \label{fig:RORD_car}
\end{figure*}

Additionally, for cross-dataset evaluation, we compare the performance of the proposed class-specific object removers with baselines using the RORD dataset \cite{Sagong_2022_BMVC}. RORD provides an image with no moving objects in a scene and images with various moving objects in the identical scene. We conduct object removal on the validation set images where humans or cars are the only moving objects in each image. We only use images where the removal targets cover 5--40\% of the images. Unlike experiments using the COCO and AV datasets where all target class objects are designated as removal targets, in the RORD images, target class objects that do not move in a scene are not set as removal targets. Therefore, the left car in the third-row image in Fig. \ref{fig:RORD_car} is not assigned as a removal target. We generate masks that only cover pixels of moving objects using semantic segmentation annotations. 

\begin{table*}[th!]
\centering

\caption{List of datasets comprising the assorted vehicle dataset.}
\vspace{0.5em}
\resizebox{0.80\textwidth}{!}{
\begin{tabular}{c|ccccc}
\hline
 & Cityscape & KITTI & KITTI-360 & Mapillary Vistas & STEP \\ \hline
Class & \begin{tabular}[c]{@{}c@{}}car, caravan, trailer,\\ truck, van\end{tabular} & \begin{tabular}[c]{@{}c@{}}bus, car, caravan,\\ trailer, truck\end{tabular} & \begin{tabular}[c]{@{}c@{}}car, caravan, trailer,\\ truck, van\end{tabular} & bus, car, truck & bus, car, truck \\ \hline
\begin{tabular}[c]{@{}c@{}}Num. images\\ w/ target class\end{tabular} & 2212 & 7467 & 12215 & 15070 & 2021 \\ \hline
\begin{tabular}[c]{@{}c@{}}Num. images\\ w/o target class\end{tabular} & 4149 & 6856 & 12310 & 2807 & 1211 \\ \hline
\end{tabular}
}
\label{table:data}
\end{table*}

\begin{figure*}[h!]
  \centering
  \includegraphics[width=0.95\textwidth]{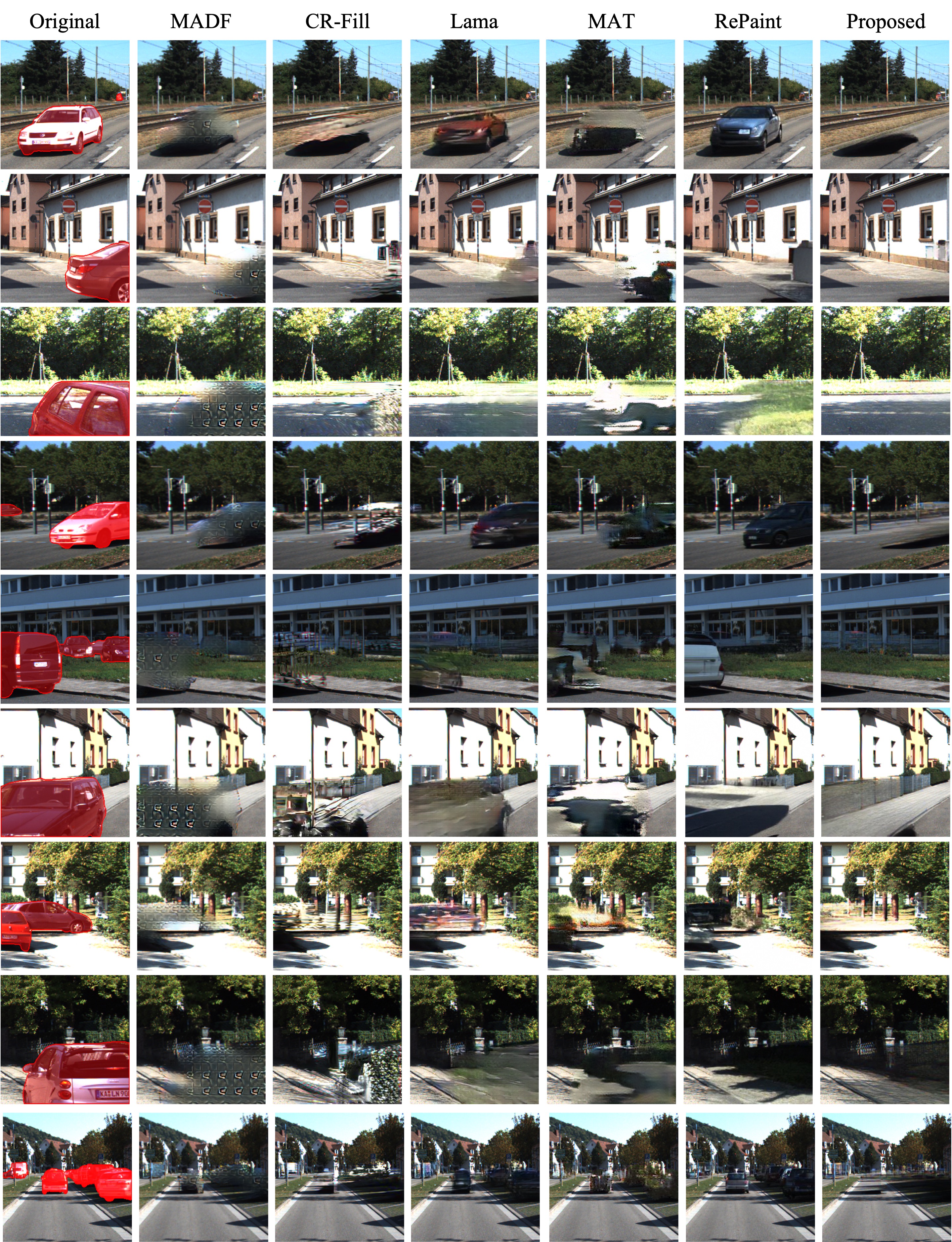}
  \caption{Object removal results on the assorted vehicle dataset. We obtain input masks for performance comparison using a semantic segmentation model (MSeg \cite{lambert2020mseg}) and use the fine-tuned models to generate the results.
  }
    \label{fig:qualitative_car}
\end{figure*}

\begin{figure*}[h!]
  \centering
  \includegraphics[width=0.95\textwidth]{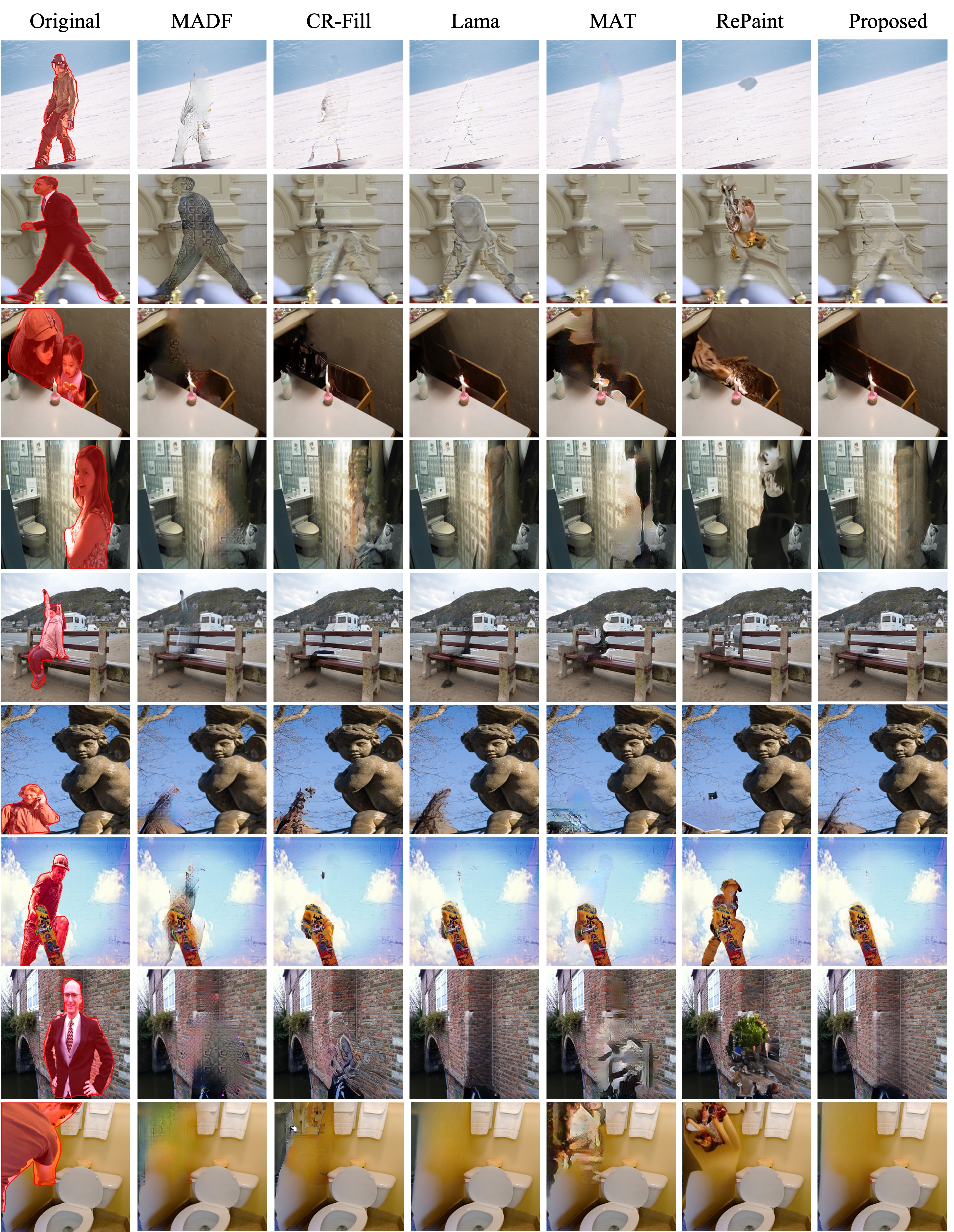}
  \caption{Object removal results on the COCO dataset. We use the ground truth semantic segmentation maps to generate input masks and use the fine-tuned models to generate the results.
  }
    \label{fig:qualitative_human}
\end{figure*}

\begin{figure*}[th!]
  \centering
  \includegraphics[width=0.98\textwidth]{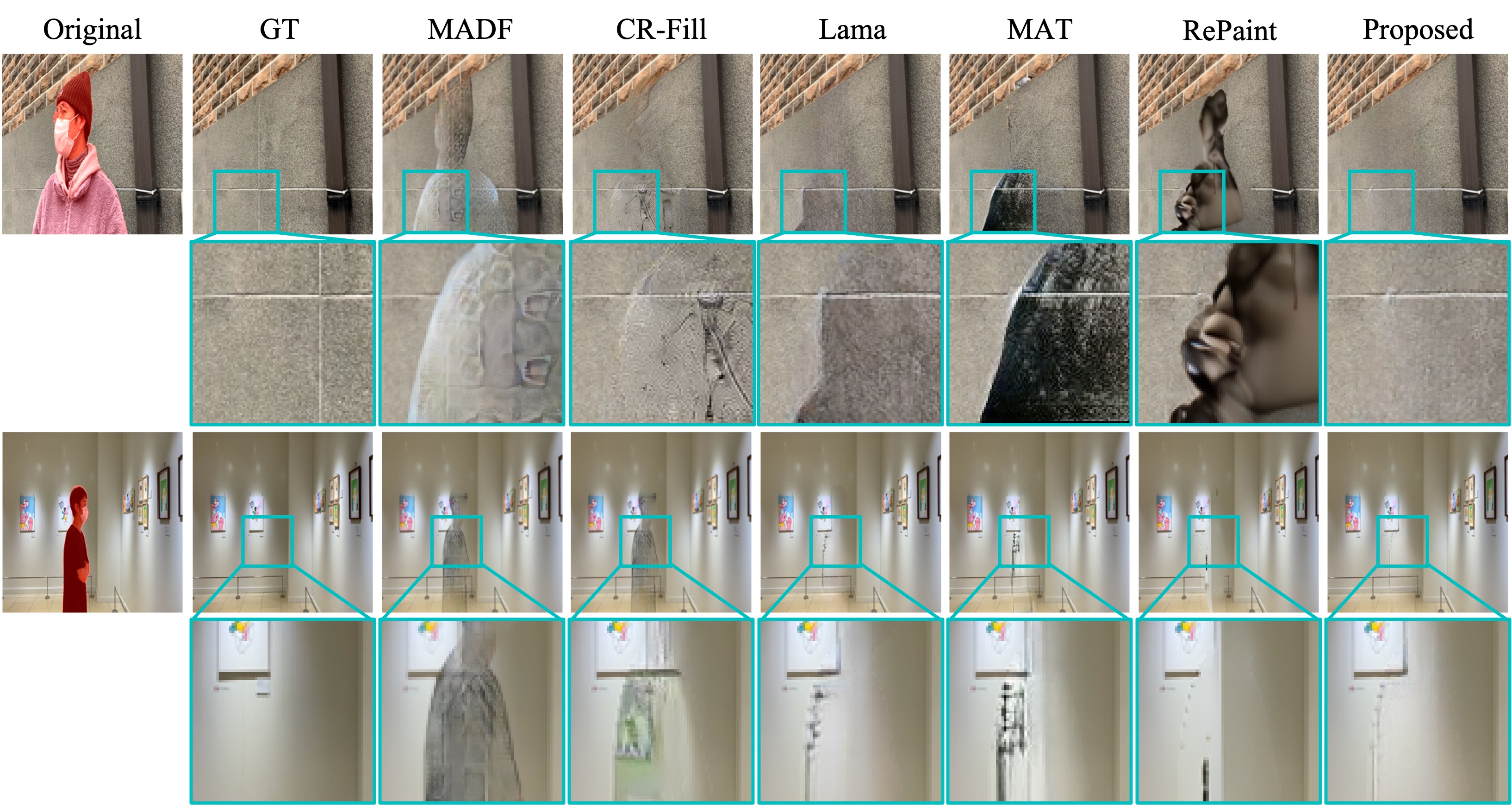}
  \caption{Human removal results on the RORD dataset. We use the ground truth semantic segmentation maps to generate input masks and use the models trained on PLACES2 to generate the results.
  }
    \label{fig:RORD_human}
\end{figure*}

\textbf{Evaluation methods}
The COCO and assorted vehicle datasets do not provide object removal GT, and it is challenging to acquire images taken under the identical environment of the images in the datasets, only without removal targets. Therefore, we employ FID$^*$ and U-IDS$^*$ \cite{oh2024object}, designed to gauge the performance of object removers without relying on object removal GT images, for performance evaluation. Unlike FID and U-IDS which do not place any condition on the images that make up a query set and comparison set, FID$^*$ and U-IDS$^*$ are calculated using class-wise object removal results as a query set and images without a target class object as a comparison set. Therefore, to exploit the evaluation methods, we set all target class objects in the image as removal targets and use test set images without target class objects as the comparison set. 

In experiments using the RORD dataset, where the images without a moving object serve as object removal GT,  we evaluate the object removal quality with full-reference image quality assessment methods that rely on reference images for evaluation, such as PSNR, SSIM, and LPIPS. We also utilize FID for performance evaluation, which uses object removal GT images as a comparison set.

\textbf{Implementation Details. }
Considering the limited size of our training datasets, we use the parameters of the backbone model trained on PLACES2 \cite{zhou2017places}(about 10 million images) as initial values, and train them on a target dataset for 50 epochs. All discriminator and generator models are trained using Adam \cite{kingma2014adam} optimizer, with a fixed learning rate of 0.001 and 0.00001, respectively. We use the NVIDIA RTX A6000 GPU for training. 

\textbf{Baselines. }
We compare our method with strong baselines, as shown in Figs. \ref{fig:qualitative_car} and \ref{fig:qualitative_human}, and Table \ref{table:quantitative}. For a fair comparison, we not only evaluate the performance of publicly available models (trained on PLACES2 \cite{zhou2017places}) but also evaluate the performance of models which are fine-tuned for 50 epochs. We follow the training process provided by the authors. For RePaint \cite{lugmayr2022repaint}, we train the DDPM model \cite{nichol2021improved} on the target dataset for 1000K iterations (batch size = 6) and utilize it as a fine-tuned model. For the sampling process of RePaint, we use 250 timesteps for sampling and apply 5 times resampling with a jump size of 5.

\begin{table*}[]
\centering
\caption{Object removal performance of baselines and the proposed method. We report FID$^*$ and U-IDS$^*$ (in \%) results on the COCO and assorted vehicle datasets. We set the \textit{person} and \textit{car} classes as target classes, repectively.  $\dagger$ indicates the fine-tuned model. The best results are underlined with bold letters.
}
\begin{tabular}{c|c|ccccccccccc}
\hline
\multirow{2}{*}{Dataset}                  & \multirow{2}{*}{\begin{tabular}[c]{@{}c@{}}Evaluation \\ metric\end{tabular}} & \multicolumn{11}{c}{Object remover}                                                                \\ \cline{3-13} 
                                          &                                    & MADF  & MADF$^{\dagger}$  & CR-Fill & CR-Fill$^{\dagger}$   & Lama  & Lama$^{\dagger}$  & MAT   & MAT$^{\dagger}$   & RePaint & RePaint$^{\dagger}$ & Proposed \\ \hline
\multirow{2}{*}{\begin{tabular}[c]{@{}c@{}}Assorted\\ vehicle\end{tabular}} & FID$^*$  $\downarrow$                                & 66.40 & 66.85 & 51.34   & 49.18     & 48.15 & 49.09 & 56.22 & 48.96 & 53.72   & 50.24   & \underline{\textbf{42.64}}    \\
                                          &  U-IDS$^*$  $\uparrow$                              & 0.00  & 0.00  & 0.24    & 0.24      & 0.89  & 0.42  & 0.00  & 0.02  & 0.29    & 0.70    & \underline{\textbf{2.61}}     \\ \hline
\multirow{2}{*}{COCO}             & FID$^*$  $\downarrow$                                & 57.10 & 56.49 & 51.89   & 52.78     & 50.72 & 50.66 & 51.62 & 54.94 & 50.16   & 51.59   & \underline{\textbf{48.87}}    \\
                                          &  U-IDS$^*$  $\uparrow$                              & 5.44  & 5.57  & 9.38    & 9.33 & 10.84 & 10.84 & 10.51 & 6.14  & 10.39   & 8.07    & \underline{\textbf{11.11}}    \\ \hline
\end{tabular}
\label{table:quantitative}
\end{table*}

\begin{table*}[]
\centering
\caption{Object removal performance of baselines and the proposed method. We report FID, LPIPS, PSNR, and SSIM results on the RORD dataset. $\dagger$ indicates the fine-tuned model. The fine-tuned models for car and human class removals are trained on the assorted vehicle and COCO datasets, respectively. The best results are underlined with bold letters.
}

\begin{tabular}{c|c|ccccccccccc}
\hline
\multirow{2}{*}{\begin{tabular}[c]{@{}c@{}}Removal target\\ class\end{tabular}} & \multirow{2}{*}{\begin{tabular}[c]{@{}c@{}}Evaluation \\ metric\end{tabular}} & \multicolumn{11}{c}{Object remover} \\ \cline{3-13} 
 &  & MADF & MADF$^{\dagger}$ & CR-Fill & CR-Fill$^{\dagger}$ & Lama & Lama$^{\dagger}$ & MAT & MAT$^{\dagger}$ & RePaint & RePaint$^{\dagger}$ & Proposed \\ \hline
\multirow{4}{*}{Car} & FID $\downarrow$ & 140.27 & 118.83 & 75.26 & 82.00 & 65.78 & 71.18 & 104.50 & 94.65 & 100.57 & 93.78 & \underline{\textbf{63.71}} \\
 & LPIPS $\downarrow$ & 0.147 & 0.141 & 0.083 & 0.084 & 0.080& 0.081 & 0.106 & 0.097 & 0.092 & 0.096 & \underline{\textbf{0.075}} \\
 & PSNR $\uparrow$ & 31.59 & 31.60 & 32.37 & 32.36 & 32.42 & 32.43 & 32.38 & 32.41 & 32.30 & 32.24 & \underline{\textbf{32.47}} \\
 & SSIM $\uparrow$ & 0.848 & 0.850 & 0.881 & 0.879 & 0.884 & 0.884 & 0.870 & 0.875 & 0.878 & 0.874 & \underline{\textbf{0.888}} \\ \hline
\multirow{4}{*}{Human} & FID $\downarrow$ & 132.63 & 108.58 & 52.51 & 55.83 & 47.06 & 46.11 & 72.47 & 55.95 & 44.84 & 47.93 & \underline{\textbf{43.84}} \\
 & LPIPS $\downarrow$ & 0.157 & 0.147 & 0.077 & 0.080 & 0.077 & 0.074 & 0.105 & 0.089 & 0.076 & 0.081 & \underline{\textbf{0.071}} \\
 & PSNR $\uparrow$ & 30.74 & 30.75 & 31.62 & 31.60 & 31.64 & 31.65 & 31.59 & 31.62 & 31.64 & 31.62 & \underline{\textbf{31.67}} \\
 & SSIM $\uparrow$ & 0.834 & 0.839 & 0.877 & 0.876 & 0.877 & 0.879 & 0.864 & 0.870 & 0.876 & 0.873 & \underline{\textbf{0.881}} \\ \hline
\end{tabular}
\label{table:RORD}
\end{table*}

\subsection{Qualitative Result}
Figs. \ref{fig:qualitative_car} -- \ref{fig:RORD_human} show the qualitative results. We can confirm that the proposed method removes target class objects in the most visually plausible way. In the baseline results in Figs. \ref{fig:qualitative_car} and \ref{fig:RORD_car}, we can easily find afterimages of the removal targets, while in the results generated by the proposed method, it is difficult to find traces of the removal targets except for the shadows. In Figs. \ref{fig:qualitative_human} and \ref{fig:RORD_human}, we can find that the outlines of the removal targets remain in all results. We infer that some pixels of the removal targets are left unmasked due to the annotation error, and these pixels are utilized during the inpainting process. Even in such situations, the proposed method fills the areas where the removal targets were located in a plausible way. 

\subsection{Quantitative Result }
The proposed class-specific object removers are evaluated to produce the best object removal results from \textit{all evaluation metrics} in experiments conducted on the assorted vehicle and COCO datasets, as shown in Table \ref{table:quantitative}. This demonstrates that the proposed class-specific object remover better erase target class objects compared to the other object removers. Lama, fine-tuned Lama (Lama$^{\dagger}$), and the proposed method use the identical model structure for removal, but it can be observed that performance significantly improves when using the proposed framework for training. This highlights that simultaneously teaching restoration and removal is one of the reasons why current image inpainting networks make unsatisfactory object removal images. We confirm that the proposed method, which separately trains these two tasks, can contribute to the enhancement of object removal performance of the inpainting model. 

In the experiment using the assorted vehicle dataset, two fine-tuned models (MADF$^{\dagger}$ and Lama$^{\dagger}$) exhibit lower object removal performance than their original model, and in the experiment using the COCO dataset, three fine-tuned models (CR-Fill$^{\dagger}$, MAT$^{\dagger}$, and RePaint$^{\dagger}$) show lower object removal performance than their original model. Since more than half of the images in the COCO and assorted vehicle datasets contain \textit{person} and \textit{car} class objects, respectively, without a tailored training process for object removal tasks, an inpainting model may perform multiple restoration tasks during fine-tuning. For this reason, we infer that in some models, additional training on the target dataset has led to a decrease in the object removal performance of the models. This result once again demonstrates that the current image inpainting training approach is not suitable for training object removal.

To compare the cross-dataset performance of baselines trained on PLACE2 and baselines fine-tuned on assorted vehicle and COCO datasets with the proposed class-specific object removers, we utilize the RORD dataset. Table \ref{table:RORD} shows the performance of the object removers tested on the RORD dataset. We can confirm that the proposed class-specific object removers are rated as the best for removing target class objects by all evaluation metrics. This result demonstrates that the proposed class-specific object remover can better remove target class objects even in images collected from different sources compared to the training images. 

\subsection{Ablation Study}

\begin{table}[t!]
\centering
\caption{Ablation study on the guidance from a class-specific object restorer and the data curation method. We report FID$^*$ and U-IDS$^*$ (in \%) results on the assorted vehicle dataset. To investigate the impact of the object restorer’s guidance on the performance of the object remover, the discriminator does not use the output of the restorer when $\mathscr{L}_{adv}$ is not used.}
\begin{tabular}{ccc|cc}
\hline
Data curation & $\mathscr{L}_{adv}$ & $\mathscr{L}_{afterimage}$ & FID$^*$ $\downarrow$  & U-IDS$^*$ $\uparrow$ \\ \hline
-             & -           & -             & 49.09 (-) & 0.42 (-)   \\
\checkmark    & -           & -             & 45.57 (+3.52) & 1.60 (+1.18) \\
-             & \checkmark  & \checkmark    & 55.49 (-6.40) & 0.40 (-0.02)  \\
\checkmark    & -           & \checkmark    & 43.52 (+5.57) & 2.03 (+1.61)  \\
\checkmark    & \checkmark  & -             & 45.28 (+3.81) & 1.93 (+1.51) \\
\checkmark    & \checkmark  & \checkmark    & 42.64 (+6.45) & 2.61 (+2.19)  \\ \hline
\end{tabular}
\label{table:ablation}
\end{table}

\textbf{Data curation. }
When comparing the performance of the first and the second rows of Table \ref{table:ablation}, we can confirm that the proposed data curation method alone significantly improves the target class object removal performance of the image inpainting network. This result demonstrates that training without the samples whose original images contain target class objects helps enhance the target class object removal performance of an image inpainting network. 

When training an inpainter using our framework without data curation, the object removal performance decreases significantly, which can be seen in the third row of Table \ref{table:ablation}. Without data curation, the training process of our framework includes both restoration and removal tasks. In this scenario, the object restorer hinders the effective restoration during training and degrades overall inpainting performance of the object remover, resulting in a decrease in the object removal performance of the object remover. Therefore, without data curation, the object removal performance of the object remover obtained using the proposed framework is lower than that of the inpainter obtained with the conventional inpainting framework.

\textbf{Guidance. }
When comparing the fourth and fifth rows of Table \ref{table:ablation} with the second row, it is evident that both guidances generated using the class-specific object restorer contribute to the performance enhancement of the class-specific object remover. In particular, we find that the guidance using $\mathscr{L}_{afterimage}$ can significantly contribute to the performance improvement of the class-specific object remover.

\section{Conclusion}
In this paper, we systematically investigate the reason behind the unsatisfactory object removal results generated by image inpainting networks. We find that the current training approaches which encourages a single inpainting model to handle both object removal and restoration tasks is one of the reasons. To address this issue, we propose the task-decoupled image inpainting framework which generates two separate inpainting models. With curated samples and training schemes tailored for training a class-specific object remover, our method can better remove target class objects compared to object removers based on image inpainting networks.

{\small
\bibliographystyle{IEEEtran}
\bibliography{bib_list}
}

\begin{comment}
\begin{IEEEbiography}[{\includegraphics[width=1in,height=1.25in,clip,keepaspectratio]{author_img/changsuk.jpg}}]{Changsuk Oh} 
received the B.S. and M.S. degrees in the Department of Mechanical and Aerospace Engineering from Seoul National University, Seoul, South Korea, in 2019 and 2021, respectively, where he is currently pursuing the Ph.D. degree in the Aerospace Engineering. His current research interest include deep learning and computer vision. 
\end{IEEEbiography}

\begin{IEEEbiography}[{\includegraphics[width=1in,height=1.25in,clip,keepaspectratio]{author_img/hjinkim.jpeg}}]{H. Jin Kim} 
received the B.S.
degree from the Korea Advanced Institute of Technology (KAIST), in 1995, and the M.S. and Ph.D.
degrees in mechanical engineering from the University of California at Berkeley (UC Berkeley), in
1999 and 2001, respectively. From 2002 to 2004,
she was a Postdoctoral Researcher of electrical engineering and computer science with UC
Berkeley. In 2004, she joined the Department of
Mechanical and Aerospace Engineering, Seoul
National University, as an Assistant Professor, where she is currently a Professor. Her research interests include intelligent control of robotic systems
and motion planning.
\end{IEEEbiography}
\end{comment}

\end{document}